\documentclass[11pt]{article}

\usepackage[margin=1in]{geometry}

\usepackage{booktabs}
\usepackage{threeparttable}
\usepackage{multicol}
\usepackage{multirow}
\usepackage{graphicx}
\usepackage{xcolor}
\usepackage{array}

\usepackage{listings}
\usepackage{xcolor}
\usepackage{courier}
\usepackage{amsmath, amssymb}
\usepackage{graphicx}
\usepackage{color}
\usepackage{hyperref}
\hypersetup{
    colorlinks=true,
    linkcolor=blue,
    citecolor=blue,
    urlcolor=blue
}
\usepackage{authblk}

\title{\bfseries CT-Agent: A Multimodal-LLM Agent for 3D CT Radiology Question Answering}
\author[1]{Yuren Mao\thanks{Contributing authors. Emails: yuren.mao@zju.edu.cn; xuwenyi@zju.edu.cn, yuyangqin@zju.edu.cn}}
\author[1]{Wenyi Xu$^*$}
\author[1]{Yuyang Qin$^*$}
\author[1]{Yunjun Gao\thanks{Corresponding author. Email: gaoyj@zju.edu.cn}}
\affil[1]{School of Software Technology, Zhejiang University, Hanghou, 310007}

\date{}  % 不显示日期

\begin{document}
\maketitle

\begin{abstract}
Computed Tomography (CT) scan, 
%which is a kind of 3D medical imaging containing hundreds of cross-sectional images (a.k.a. slices), provides detailed information for diagnosis. 
which produces 3D volumetric medical data that can be viewed as hundreds of cross-sectional images (a.k.a. slices), provides detailed anatomical information for diagnosis.
For radiologists, creating CT radiology reports is time-consuming and error-prone. A visual question answering (VQA) system that can answer radiologists' questions about some anatomical regions on the CT scan and even automatically generate a radiology report is urgently needed. However, existing VQA systems cannot adequately handle the CT radiology question answering (CTQA) task for: (1) anatomic complexity makes CT images difficult to understand; (2) spatial relationship across hundreds slices is difficult to capture. To address these issues, this paper proposes CT-Agent, a multimodal agentic framework for CTQA. CT-Agent adopts anatomically independent tools to break down the anatomic complexity; furthermore, it efficiently captures the across-slice spatial relationship with a global-local token compression strategy. Experimental results on two 3D chest CT datasets, CT-RATE and RadGenome-ChestCT, verify the superior performance of CT-Agent.

\vspace{1em}

\end{abstract}

\section{Introduction}

Computed Tomography (CT), which is a  three-dimensional X-ray-based  medical imaging technique, is widely used in the diagnosis of various diseases, such as tumors, fractures, and lung diseases \cite{mccollough2015computed, bernsen2020ct}. CT involves volumetric data, which provides a more complete view of the three-dimensional structure of a lesion and its spatial relationship with surrounding tissues compared to 2D medicine images. To process the volumetric data, CT volumes are typically reformatted into hundreds of cross-sectional slices\cite{khlaut2025radsam, cohen2025explaining, yu2024enhancing}. It puts challenges on the analysis and writing radiology reports for radiologists.

In the process of writing radiology reports, radiologists have to examine hundreds of slices one by one to identify potential abnormalities. It is time-consuming, labor-intensive, and prone to human error. There is a growing need for intelligent tools to provide auxiliary information for radiologists to accelerate the writing process and improve  quality of the reports.  A visual question answering (VQA) system that can answer radiologists’
questions about some anatomical regions on the CT scan and even automatically generate a radiology report is desired.

However, existing VQA methods\cite{li2023llava,  zhang2023pmc} cannot meet the need of question answering for CT volumes. Although recent VQA methods have made a good progress, they mainly focus on 2D images or videos. Building VQA systems for 3D CT volumetric data remains a major challenge. In CT radiology question answering (CTQA), it is necessary to accurately model anatomical structures and capture spatial relationship across slices, which presents two main challenges. First, anatomical structures are highly complex. Boundaries between organs are often unclear and their shapes vary widely. This makes it difficult for CTQA models to understand semantics and localize abnormalities. Secondly, the spatial relationship across slices is difficult to capture. Many lesions span multiple regions and slices. Existing methods\cite{wang2023r2gengpt} that rely on single images or sparse sampling struggle to build a consistent global representation. In addition, CT scans contain hundreds of slices. Encoding them directly leads to an overwhelming number of visual tokens, far beyond what current multimodal models can handle.

To address the above challenges, we propose CT-Agent, a multimodal large language models (LLMs)-driven agentic framework for CTQA. CT-Agent is anatomy-aware and token-efficient. It tackles the anatomical complexity based on an \textit{organ-specific sub-model ensemble strategy}, where we train a LoRA plugin for each anatomical region to obtain fine-grained and localized clinical clues.  Furthermore, CT-Agent captures the spatial relationship and reduces the number of input tokens by adopting a \textit{global-local dual-path token compression method}, which reduces the length of the token by approximately 75\% while preserving semantic integrity. Based on the above methods, we design an action space for CT-Agent. Facilitating by the planning ability of LLMs, CT-Agent dynamically identifies task types, dispatches reasoning paths, incorporates memory-based exemplar retrieval, and takes actions to generate coherent and clinically stylized responses. Experimental results on the two public 3D chest CT datasets (CT-RATE and RadGenome-ChestCT) demonstrate that CT-Agent consistently outperforms existing methods on both report generation and question answering tasks, showcasing its overall advantages in semantic fluency and clinical efficacy.

\section{Related Work}

Medical VQA has seen significant progress in recent years. Early methods mainly relied on template retrieval or Transformer-based sequence generation models, such as R2Gen\cite{chen2020generating} and BPI-MVQA\cite{liu2022bpi}, primarily focusing on 2D image types like chest X-rays. With the rise of large language models (LLMs) in the medical domain, approaches such as MAIRA\cite{hyland2023maira}, XrayGPT\cite{thawkar2023xraygpt}, and PMC-VQA\cite{zhang2023pmc} introduced instruction-tuned generation to enable more flexible free-text outputs. However, these methods are still limited to 2D inputs and lack effective modeling of spatial relationship and anatomical semantics inherent in 3D volumetric data.

Recent studies on 3D CT report generation have begun to move beyond traditional 2D modeling paradigms, advancing toward volumetric representations with stronger spatial reasoning capabilities. 
CT2Rep\cite{hamamci2024ct2rep} introduced a hierarchical memory mechanism to enable global encoding of volumetric data. Models such as 3D-CT-GPT\cite{chen20243d} and ViT3D\cite{li2024vit3d} integrate 3D vision encoders with large language models, achieving notable progress in cross-modal semantic understanding. 
CT-AGRG\cite{di2024ct} adopts an abnormality-centric generation strategy, enhancing clinical specificity, while Argus\cite{liu2024benchmarking} establishes a high-resolution 3D dataset to improve fidelity to real anatomical detail. 
In parallel, MS-VLM\cite{lee2024read} and M3D\cite{bai2024m3d} simulate the slice-by-slice review process of radiologists, effectively capturing contextual dependencies across slices. 
These methods have laid a solid foundation for 3D report generation, although limitations remain in integrating regional semantics and improving task generalization.

% compared 

The role of LLMs in 3D medical imaging is further explored through the development of agent-based systems. These agents, designed to automate and enhance the interaction with clinical data, are increasingly used for tasks such as radiology report generation and answering diagnostic queries\cite{wang2025survey}. 
Representative works such as M³Builder, MedAgent-Pro, and PathFinder exemplify this trend from different perspectives\cite{ghezloo2025pathfinder,wang2025medagent,feng2025m}. M³Builder streamlines the end-to-end medical machine learning workflow through coordinated multi-agent collaboration; MedAgent-Pro establishes a reasoning-centered diagnostic pipeline grounded in multimodal clinical evidence; and PathFinder emulates the decision-making process of pathologists by integrating multiple agents to analyze whole-slide images with interpretability.
Despite these advances, existing models often treat volume data as a homogeneous entity, lacking anatomical decomposition and efficient token handling. Our proposed CT-Agent addresses these limitations through anatomy-aware reasoning and a dual-path compression strategy, enabling scalable and interpretable inference.

Moreover, the advancement of medical visual question answering and report generation tasks has been strongly supported by the development of diverse and well-annotated imaging datasets. In the 2D domain, resources such as VQA-RAD\cite{liu2024pefomed}, MedMNIST\cite{medmnistv1}, and MedSegBench\cite{Ku2024} have offered rich collections of X-ray, ultrasound, and pathological images paired with structured labels or question-answer pairs, enabling foundational research in image-language understanding. On the 3D side, datasets like CT-RATE\cite{ct-rate} and RadGenome-ChestCT\cite{radgenome} have been instrumental, providing large-scale chest CT volumes with detailed radiology reports, anatomical region masks, and millions of aligned question-answer pairs. Additionally, datasets such as MedMNISTV2\cite{medmnistv2} and MedShapeNet\cite{li2025medshapenet} contribute further by offering structured 3D imaging data with clinical annotations, facilitating spatially-aware modeling of volumetric medical data. CT-Agent is built upon two representative 3D chest CT datasets, CT-RATE\cite{ct-rate} and RadGenome-ChestCT\cite{radgenome}, which provide large-scale volumetric data with region-level annotations and structured QA supervision, forming the foundation for anatomical reasoning and task-specific training.

\section{Preliminaries}

In this section, we first define the problem of 3D CT Radiology Question Answering and then introduce the 3D CT data preprocessing procedure and the background of the Low-Rank Adaptation.

\subsection{Problem Definition}

3D CT Radiology Question Answering (CTQA) aims to generate clinically meaningful answers $A$ for a user-issued natural language query $Q$ and a volumetric CT scan $I $. Formally, the objective of CTQA is to learn a multimodal mapping function:
\begin{equation}
f: (Q, I) \rightarrow A
\end{equation}
Typically, CTQA systems need to support two functional modes: 
(1) Radiology Report Generation, where the system should conduct a comprehensive analysis over the entire CT volume and provide a radiology report. 
(2) Region-Guided Question Answering, where the system should answer users' questions about one or more specific anatomical regions on a CT scan; 
The former emphasizes global contextual integration and summarization across multiple regions, while the latter demands localized semantic understanding and targeted reasoning. 
The output of CTQA should satisfy two evaluation criteria: linguistic quality and clinical correctness. 
Linguistic quality, which measures the fluency, coherence, and lexical similarity of the generated text with respect to ground truth, can be measured by BLEU, ROUGE-L, and METEOR. 
By contrast, clinical correctness measures the factual accuracy and clinical relevance of the output. It can be evaluated by Clinical Efficacy (CE), which considers the presence and correctness of medically significant entities and relations in the generated output.

\subsection{3D CT Preprocessing}
To prepare the 3D CT volumes for multimodal learning, we apply a series of preprocessing steps to ensure spatial consistency, standardized resolution, and anatomical focus\cite{hamamci2024ct2rep}. Each volumetric scan is first resampled to a uniform voxel spacing (typically 1.5mm along the z-axis and 1.0mm in the axial plane) to reduce variation across scans and facilitate consistent modeling. The spatial orientation of each scan is aligned to a canonical coordinate system, and irrelevant background regions are removed by cropping to the foreground area. Intensity values are converted into physical Hounsfield Units (HU) using the rescale parameters in the DICOM metadata, enabling a consistent representation of tissue densities\cite{bernsen2020ct}.

Following spatial and intensity normalization, each 3D CT volume is decomposed into a sequence of axial slices. To extract semantically rich visual features, we encode each slice independently using a pretrained vision transformer backbone, such as CLIP ViT-B/16\cite{radford2021learning}. Each slice is partitioned into non-overlapping patches and mapped into a high-dimensional token space, producing a structured representation that preserves local anatomical detail. The full volume is thus represented as a sequence of slice-level token matrices, which serve as input to the CT-Agent's action tools reasoning modules.

\subsection{Low-Rank Adaptation}

To efficiently adapt LLMs to domain-specific tasks such as CT radiology reasoning, full fine-tuning is often prohibitively expensive in terms of computation and memory. To address this, parameter-efficient fine-tuning (PEFT) techniques have been developed, which update only a small subset of parameters while keeping the pre-trained model weights fixed. Among these techniques, Low-Rank Adaptation (LoRA)\cite{hu2022lora} has emerged as one of the most widely recognized and effective methods, achieving competitive performance across various tasks while significantly reducing resource requirements\cite{mao2025survey}.

LoRA modifies the weight update process by constraining the updates to a low-rank decomposition. Specifically, for a pre-trained weight matrix $W_0 \in \mathbb{R}^{d_1 \times d_2}$, LoRA represents the update $\Delta W$ as a low-rank product $\Delta W = BA$, where $B \in \mathbb{R}^{d_1 \times d_r}$ and $A \in \mathbb{R}^{d_r \times d_2}$, with $d_r \ll \min(d_1, d_2)$. The forward pass is then adjusted as:$h' = W_0 x + \Delta W x = W_0 x + B (A x),$ where $x$ is the input. During training, $ W_0$ remains frozen, and only the low-rank matrices $B$ and $A$ are updated, which significantly reduces the number of trainable parameters and overall computational costs. The trained LoRA weights can either be merged with the original pre-trained weights or used independently as a plug-and-play module during inference.  In our framework, we leverage LoRA to construct a set of lightweight, region-specific adapters for anatomical reasoning. Each anatomical region (e.g., lung, heart) is equipped with a LoRA module, enabling modular fine-tuning that is both computationally efficient and anatomically specialized.

\section{Methodology}
This section introduces the overall architecture and key components of CT-Agent, including how it plans tasks, takes actions, utilizes memory.

\subsection{The Framework of CT-Agent}

\begin{figure}[t]
  \centering
  \includegraphics[width=1\textwidth]{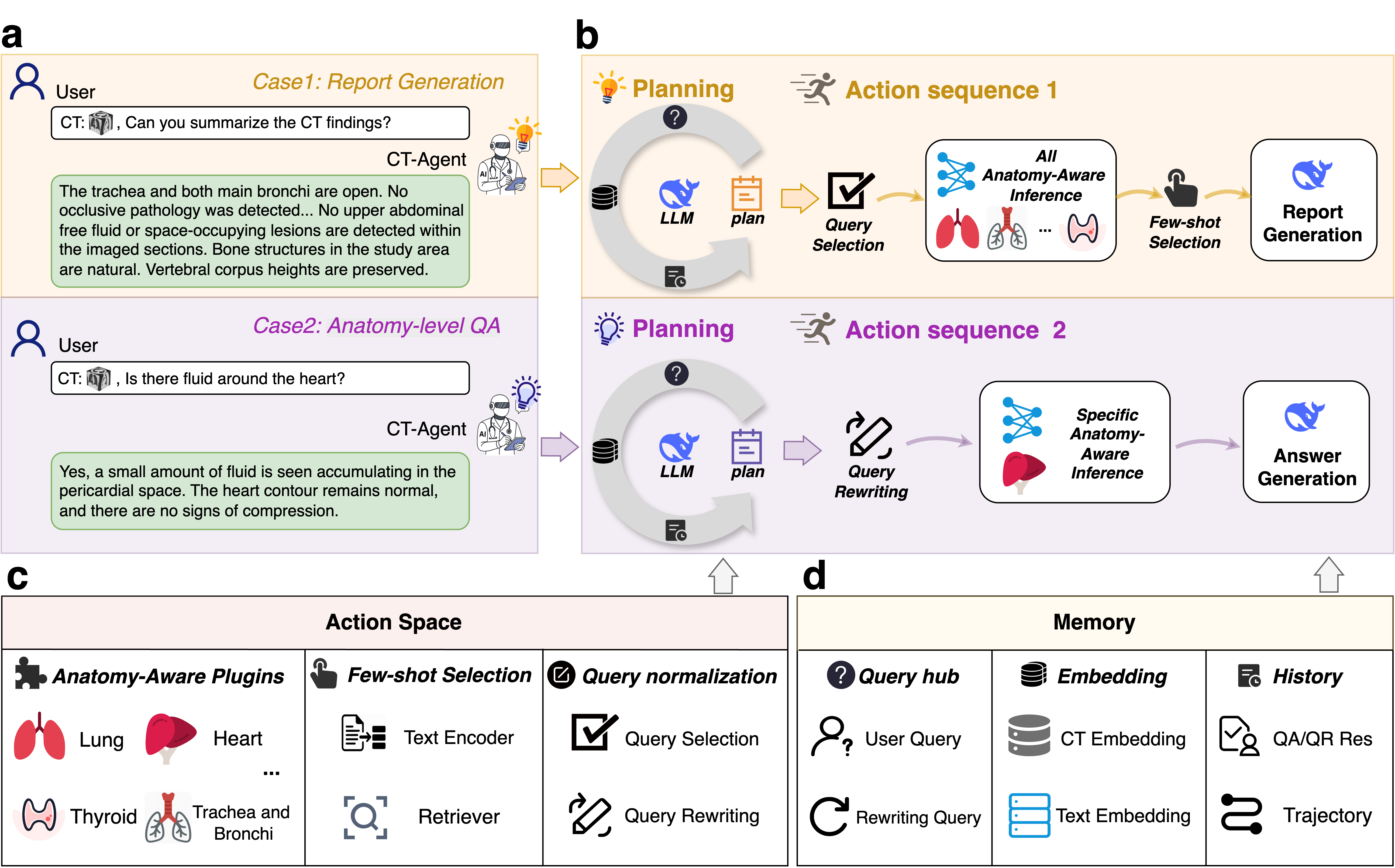}
  \caption{
    Overall architecture of CT-Agent. It consists of three modules: planning module, action space, and memory module. The planning module is driven by a LLM and is responsible for identifying the task type (Report Generation or Anatomy-level QA), parsing user inputs, locating the involved anatomical region, dispatching appropriate tool modules, and planning how to select few-shot exemplars. The action space includes a set of anatomy-aware plugins, each specialized for a different anatomical region, few-shot selection tool and query normalization tool for query selection and query rewriting. The memory module stores historical queries, planning paths, and prior radiology reports or question-answering results.
  }
  \label{fig:main}
\end{figure}

CT-Agent is an LLM agent system designed for 3D chest CT question answering and report generation. Built upon an LLM backbone, the system incorporates dynamic task planning, anatomy-aware expert tool invocation, and memory-based contextual augmentation. It is capable of performing targeted clinical reasoning based on the task type and generating outputs with semantic consistency and professional expression.

The system consists of three core components: a planning module, an action space, and a memory module. The overall architecture of CT-Agent can be formally expressed as:

\begin{equation}
\text{CT-Agent} = (\mathcal{P}, \mathcal{T}, \mathcal{M})
\end{equation}

Here, $\mathcal{P}$ denotes the planning module responsible for identifying the task type (e.g., report generation or question answering) and planning the execution path; $\mathcal{T}$ represents the action space, which includes multiple region-specific models and a few-shot selection tool; and $\mathcal{M}$ refers to the agent’s memory module.

\subsection{Planning Module}
The planning module of CT-Agent is responsible for task recognition, reasoning path planning, and tool invocation control, serving as the central scheduler of the entire system. Driven by an LLM, this module dynamically guides the downstream processing flow based on the input query and CT image. Its operational mechanism can be formalized as the following state transition function:

\begin{equation}
S_{t+1} = f(S_t, A_t, E_t)
\end{equation}

Here, $S_t$ denotes the system state at time $t$, $A_t$ represents the action taken at that moment, and $E_t$ refers to the external environment at time $t$, which includes the user-input question $Q$ and the corresponding CT volume $I$. The state transition function $f$ governs the evolution of the system by determining which tool modules to activate and whether to invoke the memory module. The action space covers operations such as task type classification, anatomical region identification, query rewriting or selection, anatomical model inference, as well as few-shot selection.

To support reasoning and decision-making, the planner internally generates structured prompts that encode the user intent and the set of available tools. These prompts are processed by the LLM, such as Deepseek-v3, to determine the appropriate downstream execution strategy. Based on the parsed semantics, the planner interacts with the action space by invoking anatomy-aligned LoRA reasoning modules and retrieval components to carry out the designated operations. As illustrated in Figure~\ref{fig:main}, the planning module supports two primary task types: radiology report generation and region-guided question answering.

In the report generation task, the planning module routes the visual input in parallel to ten region-specific reasoning tools, each responsible for localized inference within its designated anatomical area. For each region, a predefined question from a curated query pool is selected and answered by the corresponding tool to produce a diagnostic statement. These regional outputs are then aggregated by the system to form a complete radiology report. 
During the aggregation process, a memory-enhanced mechanism is employed: the planning module encodes the intermediate regional predictions into a semantic embedding and retrieves semantically similar historical exemplars from the memory module. These exemplars are incorporated into the prompt to enhance the fluency and clinical relevance of the final report.

For the region-guided question answering task, the planning module first performs query parsing to detect the anatomical focus of the user’s question. For instance, a question such as “Is there fluid around the heart?” is mapped to the heart region, as defined in the system’s ten-region anatomical taxonomy. Once the relevant region is identified, the planner invokes the query rewriting tool, which rewrites the original query into a standardized format aligned with the model’s training distribution. Next, the planner activates only the reasoning module associated with the identified region, avoiding unnecessary computation over irrelevant areas. The activated module then performs targeted inference based on both the user’s intent and the extracted visual features of that specific region.
The final answer is generated using a standardized template based on the reasoning output, ensuring clinical fluency and consistency.

\subsection{The Action Space of CT-Agent}
The action space of CT-Agent serves as the functional repository that enables the agent to perform anatomical reasoning and context-aware generation. It contains three key categories of tools: (1) anatomy-aware reasoning tools, (2) few-shot retrieval tools, and (3) query normalization tools.

\subsubsection{Anatomy-Aware Reasoning Tools}

\begin{figure}[htbp]
  \centering
  \includegraphics[width=1\textwidth]{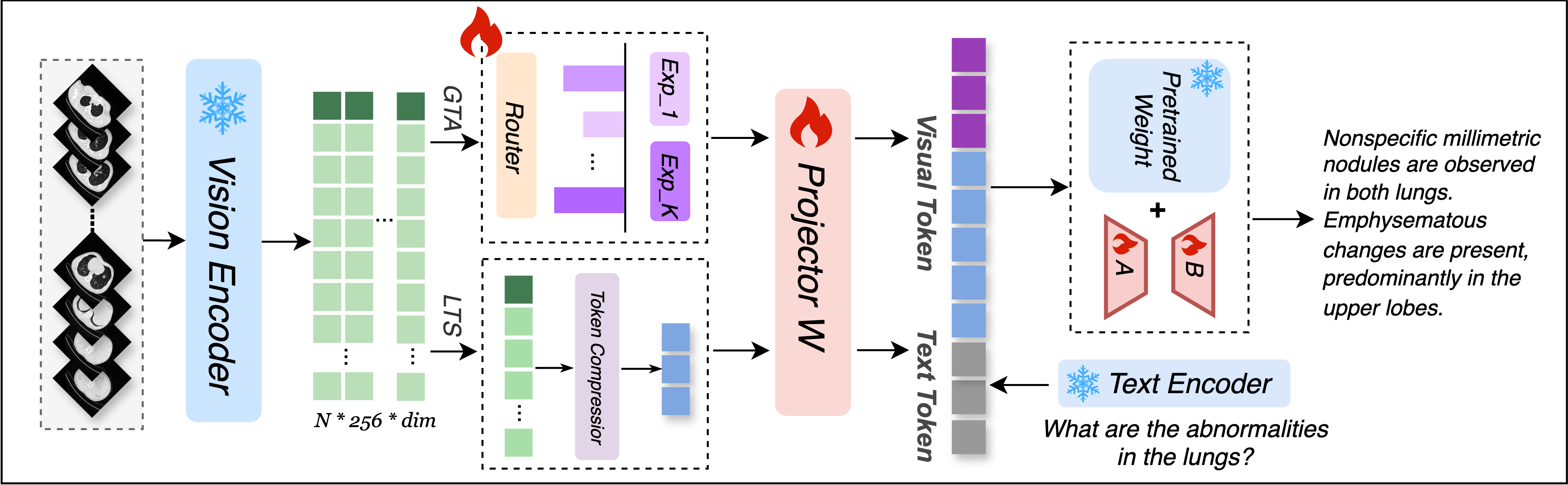}
  \caption{
    The pipeline of anatomy-aware reasoning tools. Given a set of axial CT slices, a frozen vision encoder extracts slice-level visual tokens. The token representations are processed through two parallel pathways: (1) Global Token Aggregation (GTA) and (2) Local Token Selection (LTS). The fused tokens are projected through a linear layer and combined with text tokens from the query to form a multimodal input sequence. The final response is generated by a pretrained language model augmented with LoRA plugins, enabling anatomy-specific reasoning and clinically accurate output. During training, the vision encoder and text encoder remain frozen, while the MoE module, Projector, and LoRA adapters are optimized.
  }
  \label{fig:mllm}
\end{figure}

The anatomy-aware reasoning tools consist of a collection of LoRA-weighted plugins, each specialized for a distinct anatomical region, including the lung, trachea and bronchi, mediastinum, heart, esophagus, pleura, bone, thyroid, breast, and abdomen. These models share a common multimodal LLM backbone and are selectively activated based on the task region. To ensure both computational efficiency and anatomical precision, each model is supported by a standardized processing pipeline composed of three components: hierarchical token compression, projection into the LLM embedding space, and LoRA-based training. We detail each of these components below. The overall architecture is illustrated in Figure~\ref{fig:mllm}.

\paragraph{Hierarchical Token Compression}

To alleviate the contextual burden caused by long token sequences in 3D volumes, we design a two-stage token compression consisting of a global token aggregation and a local token selection pathway.

\textbf{(1) Global Token Aggregation(GTA).} To obtain a compact semantic representation of the CT volume, we first extract slice-level visual tokens using a pretrained CLIP encoder, and then apply a token-wise Mixture-of-Experts (MoE) to the token embeddings of each slice, followed by slice-level averaging to aggregate global features.
Specifically, let the visual token embeddings extracted from the $T=240$ axial CT slices be denoted as a sequence of matrices $\{Z_t \in \mathbb{R}^{N \times d}\}_{t=1}^T$, where $N=256$ is the number of visual tokens per slice and $d=1024$ is the hidden dimension of each token. 
Each $Z_t$ represents the visual token matrix of the $t$-th slice. 
% We apply a MoE to each $Z_t$, which comprises $E=K$ expert MLPs and performs routing at the token level. 
A shared token-wise MOE is applied to each slice-level token matrix $Z_t$. It consists of $E$ independent expert MLPs, and a learned gating function assigns each input token to its top-$k$ experts, enabling conditional computation through selective expert activation.
Specifically, each token vector $z_{t,i} \in \mathbb{R}^d$ (where $i=1,\dots,N$) is routed to the top-$k$ experts according to a learned gate:
\begin{equation}
\alpha_{t,i} = \text{Softmax}(W_g z_{t,i} + b_g) \in \mathbb{R}^{E},
\end{equation}
where $W_g \in \mathbb{R}^{E \times d}$ and $b_g \in \mathbb{R}^{E}$ are learnable gating parameters. The top-$k$ entries of $\alpha_{t,i}$ (with $k \ll E$) identify the selected experts. The token is then processed by the selected experts, and the outputs are aggregated as:
\begin{equation}
z_{t,i}' = \sum_{e \in \text{TopK}(\alpha_{t,i})} \alpha_{t,i}^{(e)} \cdot \text{Expert}_e(z_{t,i}),
\end{equation}
After applying the MOE to all tokens in slice $t$, we obtain an updated matrix $Z_t' = \{z_{t,1}', z_{t,2}', \dots, z_{t,N}'\} \in \mathbb{R}^{N \times d}$, which preserves the original structure of $Z_t$ but incorporates expert-guided semantic refinement at the token level. After processing all slices through MoE, we compute the slice-wise mean to aggregate the global representation:
\begin{equation}
Z_{\text{f}} = \frac{1}{T} \sum_{t=1}^{T} Z_t' \in \mathbb{R}^{N \times d}.
\end{equation}
This produces a final global token matrix $Z_{\text{f}} \in \mathbb{R}^{256 \times 1024}$, which encapsulates the aggregated semantic content across the entire 3D volume.

\textbf{(2) Local Token Selection(LTS).} To address the intra-slice redundancy introduced by patch-based tokenization, we adopt a VisionZip-inspired\cite{yang2024visionzip} two-stage local token compression strategy consisting of dominant token selection and contextual token merging.

\textit{Dominant Token Selection.} We first identify a subset of highly informative visual tokens based on their attention interactions with the CLS token. Let $A \in \mathbb{R}^{H \times N \times N}$ be the self-attention map from a selected transformer layer in the CLIP encoder, where $H$ is the number of heads and $N$ is the number of visual tokens per slice. For each token $j$, we compute its attention score as:
\begin{equation}
\text{score}_j = \sum_{h=1}^{H} A_{h, \text{CLS}, j}, \quad j = 1, \ldots, N
\end{equation}
The top-$K$ tokens with the highest attention scores are selected as dominant tokens. Denote these selected token vectors as $\{z^{(1)}_{\text{dom}}, \ldots, z^{(K)}_{\text{dom}}\} \subset \mathbb{R}^d$.

\textit{Contextual Token Merging.} To preserve long-tail or background semantics, we compress the remaining $N - K$ tokens by merging them based on key similarity. Let the remaining tokens be split into two sets: a target set $T \in \mathbb{R}^{M \times d}$ and a merge set $M' \in \mathbb{R}^{(N-K-M) \times d}$, where $d$ is the token embedding dimension and $M$ is a predefined hyperparameter specifying the number of output contextual tokens. We calculate similarity using dot-product between the keys of merge and target tokens:
\begin{equation}
\text{sim}(m_i, t_j) = \langle K_{m_i}, K_{t_j} \rangle
\end{equation}
Each merge token $m_i$ is assigned to the most similar target token $t_j$ via $\arg\max$ operation. The assigned tokens are then aggregated using average pooling:
\begin{equation}
\text{merged}_j = \frac{1}{|\mathcal{A}_j|} \sum_{m_i \in \mathcal{A}_j} m_i, \quad \mathcal{A}_j = \left\{ m_i \mid t_j = \arg\max_{t_k \in T} \text{sim}(m_i, t_k) \right\}
\end{equation}

The final local token matrix $Z_{\text{local}} \in \mathbb{R}^{(K+M) \times d}$ is obtained by concatenating the selected dominant tokens and the generated contextual tokens:
\begin{equation}
Z_{\text{local}} = [z^{(1)}_{\text{dom}}, \ldots, z^{(K)}_{\text{dom}}, z^{(1)}_{\text{ctx}}, \ldots, z^{(M)}_{\text{ctx}}]
\end{equation}
This attention-driven compression preserves salient visual cues while significantly reducing token length, ensuring that only the most informative features are retained for efficient integration with textual queries in the anatomy-aware reasoning.

\paragraph{Projection}

To align the compressed visual representation with the input space of the LLM, we apply a linear projection to the concatenated global and local tokens. Specifically, the combined visual token sequence is first constructed as:
\begin{equation}
Z_{\text{vision}} = [Z_{\text{global}}; Z_{\text{local}}] \in \mathbb{R}^{L \times d}
\end{equation}
where $Z_{\text{global}} \in \mathbb{R}^{G \times d}$ and $Z_{\text{local}} \in \mathbb{R}^{L' \times d}$ represent the global and local token matrices respectively, $L = G + L'$, and $d = 1024$ denotes the original CLIP token dimension. To match the hidden space of the LLM, we project each token embedding from $d = 1024$ to $d' = 4096$ using a learnable linear transformation:
\begin{equation}
Z_{\text{proj}} = Z_{\text{vision}} W_p + b_p, \quad W_p \in \mathbb{R}^{1024 \times 4096},\; b_p \in \mathbb{R}^{4096}
\end{equation}
The output $Z_{\text{proj}} \in \mathbb{R}^{L \times 4096}$ serves as the visual input to the multimodal LLM, ensuring integration with text embeddings for anatomy-aware reasoning tasks.

\paragraph{Training and Inference}

CT-Agent adopts a modular training strategy to support anatomy-specific reasoning while keeping the base multimodal LLM backbone frozen. Each anatomical region $r_i$ is associated with LoRA plugin $\phi_{r_i}$, which is injected into selected layers of the transformer to enable parameter-efficient adaptation. These plugins operate within a shared vision-language model (e.g., LLaVA-Med), and are only activated during region-relevant tasks.

\textit{Training Phase.}
Given a CT volume that has been processed by the preprocessing pipeline and a region-level task prompt (either a question or a region-guided report instruction), the visual input is first transformed into a token matrix $Z_{\text{proj}} \in \mathbb{R}^{L \times d'}$ through the hierarchical token compression and projection modules described above. The task prompt and visual tokens are then concatenated to form the input sequence for the language model:
\begin{equation}
X_{\text{input}} = [T_{\text{task}}, \texttt{<im\_start>}, Z_{\text{proj}}, \texttt{<im\_end>}].
\end{equation}
Only the LoRA adapter $\phi_{r_i}$ corresponding to the annotated anatomical region is activated, while the backbone parameters $\theta$ remain frozen. The model is optimized using a standard language modeling loss 
$
\mathcal{L} = -\log P(Y \mid X_{\text{input}}; \theta, \phi_{r_i}),
$
where $Y$ is the expected output text, such as a region-specific diagnostic statement or answer.

\textit{Inference Phase.}
At inference time, the planner first parses the user query and identifies the target anatomical region. The corresponding LoRA plugin is dynamically loaded into the model. The system then encodes the preprocessed CT slides, generates $Z_{\text{proj}}$, and constructs the input sequence $X_{\text{input}}$ as in training. The LLM generates the output sequence in an autoregressive manner by predicting the next token conditioned on the previous context. Formally, given the input $X_{\text{input}}$, the model outputs the predicted text $\hat{Y} = \{\hat{y}_1, \hat{y}_2, \ldots, \hat{y}_T\}$ as:
\begin{equation}
P(\hat{Y} \mid X_{\text{input}}; \theta, \phi_{r^*}) = \prod_{t=1}^{T} P(\hat{y}_t \mid X_{\text{input}}, \hat{y}_{<t}; \theta, \phi_{r^*}),
\end{equation}
where $\hat{y}_t$ is the $t$-th predicted token, and $\phi_{r^*}$ denotes the LoRA adapter corresponding to the selected anatomical region.
This modular approach enables fine-grained specialization across regions while maintaining scalability and low adaptation overhead.

\subsubsection{Few-Shot Selection Tools}

\begin{figure}[t]
  \centering
  \includegraphics[width=1\textwidth]{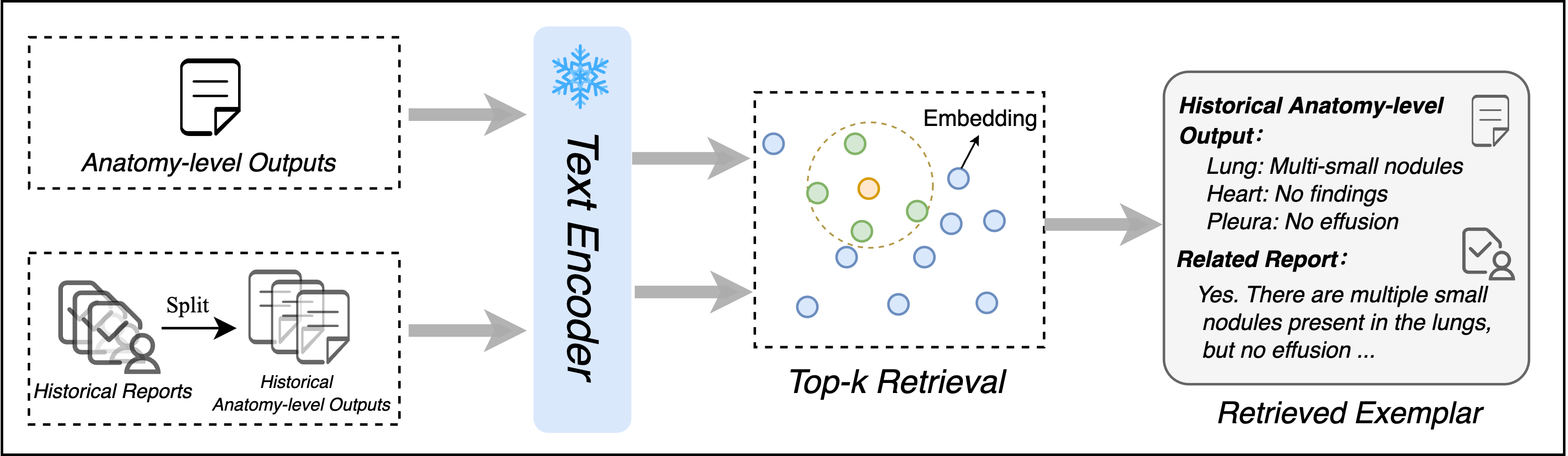}
  \caption{
    Semantic retrieval pipeline for few-shot prompting. The current case’s anatomy-level outputs are encoded into a semantic vector and matched against a vector index constructed from historical reports. Top-matching exemplars are retrieved and prepended to the input prompt for final report generation.
    }

  \label{fig:main-fig}
\end{figure}

To enhance reasoning quality in report generation, we introduce a memory-guided exemplar retrieval mechanism. This tool retrieves semantically aligned few-shot exemplars by conditioning on the model’s own anatomy-level outputs. The overall architecture is illustrated in Figure \ref{fig:main-fig}.

\textit{Encoding.}
Before encoding, the planner selects a predefined question $q_i$ for each anatomical region and invokes the corresponding reasoning tool to generate a diagnostic statement $s_i$. The resulting $s_i$ are concatenated into a sequence $s_{\text{query}} = [s_1; s_2; \dots; s_{10}]$.
We encode this sequence using a frozen sentence embedding model to obtain a semantic query vector $v_{\text{query}}$.
To support retrieval, we construct a semantic few-shot corpus $\mathcal{C} = \{(v_k, x_k)\}_{k=1}^{K}$, initialized from historical reports in the training set. Each $x_k$ is decomposed into ten anatomical findings $\{s_1^{(k)}, \dots, s_{10}^{(k)}\}$. The sequence
$s_k = [s_1^{(k)}; \dots; s_{10}^{(k)}] $
is then encoded to obtain the semantic key $v_k = \text{Encoder}(s_k)$. All $\{v_k\}$ vectors are stored in a vector index for efficient similarity search.

\textit{Retrieval.}
At inference time, cosine similarity is computed between $v_{\text{query}}$ and all stored $\{v_k\}$:
\begin{equation}
\text{sim}(v_{\text{query}}, v_k) = \frac{v_{\text{query}}^\top v_k}{\|v_{\text{query}}\| \cdot \|v_k\|}
\end{equation}
The top-$K'$ most similar exemplars $\mathcal{X}_{\text{shot}} = \{x_1, \dots, x_{K'}\}$ are retrieved and prepended to the input prompt as few-shot demonstrations. This anatomy-conditioned retrieval strategy enables CT-Agent to ground generation in clinically relevant cases, enhancing factual accuracy, stylistic fluency, and robustness to anatomical variation.

\subsubsection{Query Normalization tools}
The query normalization tools are designed to bridge the gap between free-form user questions and the query expected by anatomy-specific reasoning modules. This component includes two distinct mechanisms: query selection and query rewriting.

\textit{Query Selection.} In the report generation task, CT-Agent generates diagnostic statement for each of the ten anatomies. For this purpose, a predefined query is assigned to each anatomy in advance. These queries are designed to cover common clinical concerns and remain fixed across all report generation cases.

\textit{Query Rewriting.} In the question answering task, user-issued queries may vary widely in form and clarity. To standardize these inputs, CT-Agent employs a prompt-based rewriting method that reformulates the original question into a clear and well-structured template. The rewriting is conditioned on the identified anatomy and ensures alignment with the model’s training distribution.

\subsection{Memory Module}

Complementing the planning and action space modules, the memory module in CT-Agent serves as a persistent knowledge base for supporting anatomy-aware reasoning, few-shot selection and decision traceability. It comprises three main components: query hub, embedding store, and history log.

The query hub stores both the user-issued queries and a predefined question pool aligned with anatomical regions. For each region, the system maintains a curated set of canonical questions, which are used to guide the reasoning tools during report generation and anatomy-specific QA.

The embedding store manages semantic vector representations that support few-shot selection and vision-language alignment. It maintains two types of embeddings: (1) region-level CT embeddings, obtained by encoding axial CT slices through the vision encoder and hierarchical token compression pipeline, followed by MoE and projection; and (2) anatomy-level text embeddings derived from concat sequence using a frozen sentence encoder.

The history log records previously generated QA responses, report outputs, and reasoning trajectories during multi-step inference. This log facilitates session-level memory, supports agent interpretability, and provides context-aware grounding for future decisions.

\section{Experiments}
In this section, we perform comprehensive experiments to validate the effectiveness of our proposed CT-Agent.

\subsection{Experimental Settings}

\subsubsection{Datasets}

The experiments were conducted on two public chest CT datasets. (i) \textbf{CT-RATE}\cite{ct-rate} is a large-scale collection of 50,188 raw chest CT volumes (47,149 for training, 3,039 for testing) from 21,304 unique patients, covering a wide range of clinical cases and imaging variations. (ii) \textbf{RadGenome-Chest CT}\cite{radgenome} extends CT-RATE by providing region-wise segmentation masks for ten anatomical structures and GPT-generated pathological labels, enabling region-aligned multimodal learning. The dataset includes 25,692 non-contrast 3D chest CT scans (24,128 for training, 1,564 for testing) from 20,000 patients, 665k (624,876 for training, 40,342 for testing) multi-granularity grounded reports and 1.3M (1,343,057 for training, 84,625 for testing) grounded visual question-answering pairs. Based on the two datasets, we constructed a question-answer (QA) pair dataset for chest CT images. This dataset comprises 2,033,648 QA pairs (1,914,448 for training, 119,200 for testing), including question and answer labels for individual organs in each CT image from the CT-RATE and RadGenome-Chest CT datasets. The question templates are categorized into two primary types: \textit{presence detection} and \textit{abnormality identification}. We design unified templates for each question type and fill in the anatomical region names using the RadGenome anatomical hierarchy. For example, a presence detection question template may be “Is there any abnormality in the \texttt{\{anatomical region\}}?” The answer labels are directly derived from the GPT annotations provided by RadGenome-Chest CT, and are used as the ground-truth supervision during training. This structured design ensures both medical semantic accuracy and enables large-scale generation of question-answer pairs to support our subsequent training tasks.

\subsubsection{Baselines}

To validate the effectiveness of our proposed method, we compare it against five representative baselines for 3D medical report generation.
 (i) \textbf{CT2Rep}\cite{hamamci2024ct2rep} is the first approach specifically designed for 3D chest CT report generation. It employs an auto-regressive transformer with a memory-driven decoder and integrates longitudinal multimodal data through a hierarchical memory module. (ii) \textbf{3D-CT-GPT}\cite{chen20243d} builds upon vision––language integration by pairing a CT-specific Vision Transformer with a large language model. It uses average-pooled 3D features projected into the LLM space, supporting both VQA and report generation with strong semantic alignment. (iii) \textbf{MS-VLM}\cite{lee2024read} simulates radiologists’ slice-by-slice workflow, combining a 2D ViT, Z-former for inter-slice modeling, and a perceiver resampler to generate compact visual prompts for LLMs, enabling efficient volumetric reasoning without 3D redundancy. (iv) \textbf{M3D}\cite{bai2024m3d} blends local and global 3D features via cross-modal attention in a hierarchical encoder, enhancing lesion details and diagnostic accuracy. Together, these baselines encompass global volumetric encoding, vision–language integration, slice-level processing, and multimodal large-language modeling, providing a comprehensive foundation for evaluating our proposed approach.
 (v) \textbf{LLaVA-CT} builds on the LLaVA-Med\cite{li2023llava} by directly using the CT images and medical reports provided by CT-RATE\cite{ct-rate} for end-to-end fine-tuning.

\subsubsection{Evaluation Metrics}

To evaluate the effectiveness, we adopt two distinct evaluation frameworks for the report generation and region-guided question-answering tasks. For the report generation task, we first  employ standard natural language generation (NLG) metrics, including BLEU-1 to BLEU-4\cite{papineni2002bleu}, ROUGE-L\cite{lin-2004-rouge}, and METEOR\cite{banerjee2005meteor}. These metrics evaluate the lexical similarity and fluency of the generated reports with respect to the ground-truth reports. Second, to assess the clinical reliability of the generated content, we incorporate the Clinical Efficacy (CE) metric proposed in CT2Rep\cite{hamamci2024ct2rep}, which evaluates the model's performance at the medical semantic level by comparing the consistency of key clinical abnormalities and diagnostic terms between the generated report and the reference report. In this work, we implement the metric using an improved Exact Match (EM) method: if an abnormal term or key medical keyword that appears in the ground-truth report is accurately mentioned in the generated report, it is considered a successful match.
Specifically, we compare the set of abnormalities identified in the generated report against those annotated in the ground-truth report, focusing on the 18 common chest CT abnormalities defined in CT2Rep. Each abnormality is treated as an independent category, and we compute micro-averaged precision, recall, and F1 score based on per-abnormality agreement with the ground-truth findings. 
In the region-guided question answering task, we also adopt the Clinical Efficacy (CE) metric based on the improved EM method to evaluate the medical accuracy of generated answers across 10 anatomical regions, including the lungs, heart, trachea \& bronchi, mediastinum, pleura, bones, thyroid, breasts, esophagus, and abdomen.

\subsubsection{Implementation Details}

\textbf{3D CT Data Preprocess}. We preprocess each 3D CT volume using the MONAI framework\footnote{https://monai.io/} by aligning spatial orientation, cropping the anatomical foreground, and resampling to a fixed voxel spacing of $(1.5, 1.0, 1.0)$ mm. Voxel intensities are converted to Hounsfield Units using DICOM metadata. Each volume is then decomposed into 240 axial slices, which are resized and encoded independently using a frozen CLIP ViT-B/16 model. Each slice yields 256 patch tokens of 1024 dimensions, forming a tensor of shape $240 \times 256 \times 1024$ as visual input for the CT-Agent reasoning module.

\textbf{Tools in Action Space}. 
For the Anatomy-Aware Reasoning Tools, it is built on LLaVA-Med-v1.5~\footnote{https://huggingface.co/microsoft/llava-med-v1.5-mistral-7b}, inheriting the core parameters of its language architecture. Each LoRA plugin is injected into all attention and linear layers of the transformer backbone, with rank $r=16$, scaling factor $\alpha=16$, and dropout rate $0.05$. These adapters are implemented and managed using the PEFT library. In token compression, each CT slice is compressed into 64 visual tokens, including 54 dominant tokens and 10 contextual tokens. During training, the CLIP visual are frozen. We optimize the MoE modules, projection layers, and LoRA parameters using AdamW ($\text{lr}=2\times10^{-4}$, batch size 8) with a 2‑epoch schedule and 500‑step warm-up. Training is conducted on 8×A40 GPUs using FSDP with FP16 and gradient accumulation.
For the Few-Shot Selection Tool, we use OpenAI’s \text{text-embedding-3-small} \footnote{https://platform.openai.com/docs/models/text-embedding-3-small} to retrieve the top-3 semantically similar cases from a corpus of 20,000 pre-embedded clinical examples.

\textbf{CT-Agent Planning}. We use Deepseek-v3.0 as the backbone of the CT-Agent planning module. Task classification, anatomy identification, and other planning operations are implemented via structured prompts. All prompt templates are provided in the Appendix A.

\subsection{Overall Performance}
\paragraph{Report Generation Quality.} 
The experimental results of report generation quality are detailed in Table~\ref{tab:report-comparison}. Our method achieves state-of-the-art performance on clinical efficacy (CE) metrics and all natural language generation (NLG) metrics except BLEU-4. Specifically, compared to the strongest baseline, our approach yields improvements of 0.06 in BLEU-1, 0.029 in BLEU-2, 0.004 in BLEU-3, 0.022 in ROUGE-L, and 0.004 in METEOR, indicating enhanced fluency, coherence, and semantic alignment. In terms of clinical consistency, our method demonstrates substantial gains in CE metrics, with absolute improvements of 0.016, 0.148, and 0.159 in Precision, Recall, and F1 score, respectively. These results highlight the model’s superior capability in capturing critical medical findings accurately. As illustrated in Figure~\ref{fig:example}, we provide a comparison example between our generated report, the baseline, and the ground-truth. The performance gains can be attributed to our anatomy-aware architecture, which integrates region-specific LoRA sub-models, hierarchical token compression to preserve semantic integrity, intelligent task routing for focused reasoning, and prediction-guided exemplar retrieval to enhance contextual relevance.

\definecolor{mygreen}{RGB}{36,155,98}
\definecolor{myred}{RGB}{255,0,0}
\definecolor{myblue}{RGB}{35,125,255}
\definecolor{myy}{RGB}{255,255,102}
\begin{figure}[!t]
  \centering
  \includegraphics[width=1\textwidth]{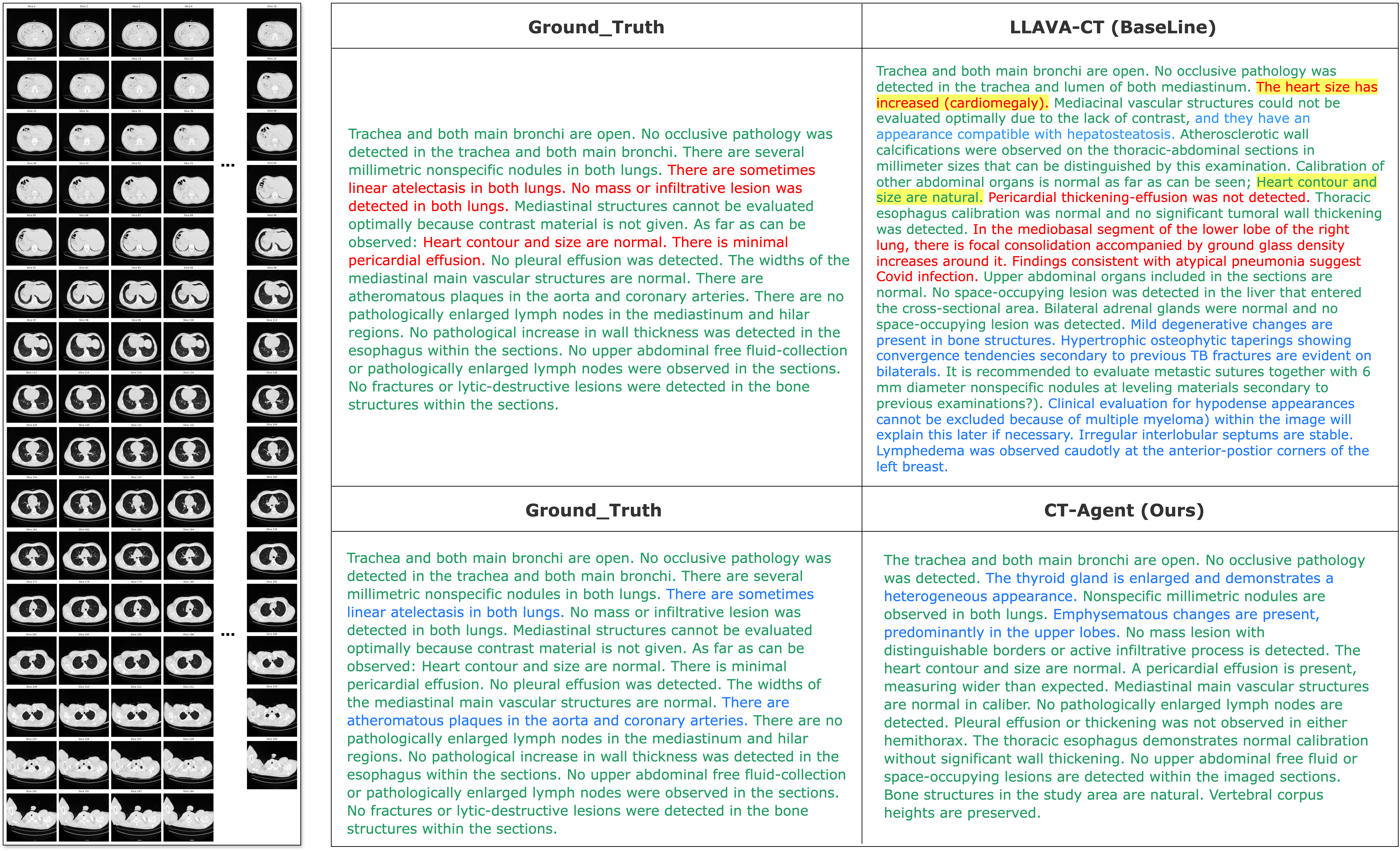}
  \caption{
    Sentence-level comparison among the CT-Agent generated report, baseline generated report and the reference report (\textit{ground\_truth}). 
    \textcolor{mygreen}{Green} highlights indicate consistent findings between the two reports. 
    \textcolor{myblue}{Blue} highlights represent mismatches or deviations: When applied to the ground truth, they mark clinically important details that the model failed to include. When appearing in the generated report, they highlight content that does not exist in the reference report, which may suggest redundancy or hallucination.
    \textcolor{myred}{Red} highlights indicate statements in the generated report that are factually incorrect or contradictory to the ground truth.
    \colorbox{myy}{Yellow background} indicates internally contradictory content within the same generated report.
  }
  \label{fig:example}
\end{figure}

\begin{table*}[ht]
\centering
\small
\caption{Performance comparison of our method with baseline models on radiology report generation.
We report standard natural language generation (NLG) metrics: BLEU-1 to BLUE-4 (BL-1 to BL-4), ROUGE-L (RL), and METEOR (M), as well as clinical efficacy metrics: Precision (P), Recall (R), and F1 score (F1) on the generated reports.}
\label{tab:report-comparison}
\begin{tabular}{
    >{\centering\arraybackslash}p{2.4 cm} |
    >{\centering\arraybackslash}p{1.09cm}
    >{\centering\arraybackslash}p{1.09cm}
    >{\centering\arraybackslash}p{1.09cm}
    >{\centering\arraybackslash}p{1.09cm}
    >{\centering\arraybackslash}p{1.09cm}
    >{\centering\arraybackslash}p{1.09cm} |
    >{\centering\arraybackslash}p{1.09cm}
    >{\centering\arraybackslash}p{1.09cm}
    >{\centering\arraybackslash}p{1.09cm}
}
\toprule
\textbf{Method} & \textbf{BL-1} & \textbf{BL-2} & \textbf{BL-3} & \textbf{BL-4} & \textbf{RL} & \textbf{M} & \textbf{P} & \textbf{R} & \textbf{F1} \\
\midrule
3D-CT-GPT     & -     & -     & -     & 0.133 & 0.145 & 0.140 & -     & -     & -     \\
CT2Rep        & 0.442 & 0.344 & 0.279 & 0.235 & 0.401 & 0.309 & 0.355 & 0.132 & 0.175 \\
MS-VLM        & -     & -     & -     & 0.232 & 0.438 & 0.396 & 0.222 & 0.329 & 0.261 \\
M3D           & 0.435 & 0.345 & 0.286 & 0.245 & 0.400 & 0.326 & 0.407 & 0.009 & 0.148 \\
LLaVA-CT      & 0.369 & 0.309 & 0.275 & \textbf{0.252} & 0.468 & 0.421 & 0.323 & 0.182 & 0.221 \\
\textbf{CT-Agent} & \textbf{0.502} & \textbf{0.374} & \textbf{0.290} & 0.231 & \textbf{0.490} & \textbf{0.425} & \textbf{0.423} & \textbf{0.477} & \textbf{0.420} \\
\bottomrule
\end{tabular}
\vspace{0.5em}
\end{table*}

\paragraph{Region-Guided Question Answering Quality.} 

To further evaluate the QA capability of our system, we measure its performance across anatomical regions on two tasks: \textit{presence detection} and \textit{abnormality identification}. As shown in Table~\ref{tab:agent-comparison}, we report P, R, and F1 scores for each anatomical region and task type. Analysis of these metrics reveals that our proposed CT-Agent consistently achieves superior performance compared to the end-to-end baseline LLaVA-CT across both evaluated tasks, with the most significant gains observed in abnormality identification. The performance of CT-Agent varies with anatomical complexity. It shows pronounced advantages in recall for high-complexity regions like the mediastinum and Trachea \& bronchi, better capturing sparse abnormal signals while maintaining robust performance in standardized regions like the Lung and Heart, reducing missed diagnoses. The improvement is more substantial for the abnormality task, where CT-Agent overcomes LLaVA-CT's tendency for complete failure in complex abnormal patterns. 
These results highlight the effectiveness of our method in accurately identifying anatomical structures and detecting abnormalities, improving its clinical utility.

\begin{table}[t]
\centering
\caption{Performance comparison between CT-Agent(denoted as O.M.) and the end-to-end baseline LLaVA-CT(denoted as BL) on visual question answering tasks across 10 anatomical regions.}
\label{tab:agent-comparison}
\footnotesize
% \resizebox{1.03\textwidth}{!}{
\setlength{\tabcolsep}{6 pt}
\renewcommand{\arraystretch}{1.1}

\begin{tabular}{
    >{\centering\arraybackslash}p{2.7 cm} |
    >{\centering\arraybackslash}p{0.69cm}
    >{\centering\arraybackslash}p{0.69cm}  |
    >{\centering\arraybackslash}p{0.69cm}
    >{\centering\arraybackslash}p{0.69cm} |
    >{\centering\arraybackslash}p{0.69cm}
    >{\centering\arraybackslash}p{0.69cm} |
    >{\centering\arraybackslash}p{0.69cm}
    >{\centering\arraybackslash}p{0.69cm} |
    >{\centering\arraybackslash}p{0.69cm}
    >{\centering\arraybackslash}p{0.69cm} |
    >{\centering\arraybackslash}p{0.69cm} 
    >{\centering\arraybackslash}p{0.69cm} 
}

\toprule
\multirow{3}{*}{\textbf{Anatomical Part}} 
& \multicolumn{6}{c}{\textbf{Presence}} 
& \multicolumn{6}{c}{\textbf{Abnormality}} \\
\cmidrule(lr){2-7} \cmidrule(lr){8-13}
& \multicolumn{2}{c}{\textbf{P}} & \multicolumn{2}{c}{\textbf{R}} & \multicolumn{2}{c}{\textbf{F1}}
& \multicolumn{2}{c}{\textbf{P}} & \multicolumn{2}{c}{\textbf{R}} & \multicolumn{2}{c}{\textbf{F1}} \\
\cmidrule(lr){2-3} \cmidrule(lr){4-5} \cmidrule(lr){6-7}
\cmidrule(lr){8-9} \cmidrule(lr){10-11} \cmidrule(lr){12-13}
& \textbf{BL} & \textbf{O.M.} &  \textbf{BL} & \textbf{O.M.} & \textbf{BL} & \textbf{O.M.} 
& \textbf{BL} & \textbf{O.M.} & \textbf{BL} & \textbf{O.M.} &  \textbf{BL} & \textbf{O.M.} \\
\midrule
Lung         & 0.973 & 0.962 & 0.904 & 0.941 & 0.937 & 0.952 & 0.934 & 0.917 & 0.362 & 0.410 & 0.521 & 0.566 \\
Trachea \& bronchi   & 0.188 & 0.145 & 0.043 & 0.349 & 0.070 & 0.205 & 0.055 & 0.887 & 0.011 & 0.713 & 0.018 & 0.790 \\
Mediastinum  & 0.720 & 0.931 & 0.663 & 0.418 & 0.690 & 0.577 & 0.633 & 0.622 & 0.446 & 0.969 & 0.523 & 0.758 \\
Heart        & 0.848 & 0.868 & 0.857 & 0.894 & 0.853 & 0.880 & 0.827 & 0.842 & 0.733 & 0.728 & 0.777 & 0.781 \\
Esophagus    & 0.279 & 0.219 & 0.241 & 0.247 & 0.259 & 0.232 & 0.276 & 0.217 & 0.165 & 0.245 & 0.207 & 0.230 \\
Pleura       & 0.676 & 0.713 & 0.232 & 0.246 & 0.345 & 0.366 & 0.598 & 0.623 & 0.182 & 0.165 & 0.279 & 0.260 \\
Bone         & 0.721 & 0.763 & 0.271 & 0.723 & 0.393 & 0.743 & 0.589 & 0.559 & 0.150 & 0.285 & 0.239 & 0.377 \\
Thyroid      & 0.981 & 0.962 & 0.830 & 0.934 & 0.899 & 0.948 & 0.953 & 0.906 & 0.377 & 0.302 & 0.540 & 0.453 \\
Breast       & 0.964 & 0.960 & 0.823 & 0.812 & 0.888 & 0.880 & 0.905 & 0.934 & 0.325 & 0.487 & 0.478 & 0.640 \\
Abdomen      & 0.933 & 0.812 & 0.396 & 0.584 & 0.556 & 0.679 & 0.882 & 0.716 & 0.214 & 0.341 & 0.345 & 0.462 \\
\midrule
\textbf{Overall} & 0.728 & 0.734 & 0.526 & 0.615 & 0.589 & 0.646 & 0.665 & 0.722 & 0.297 & 0.465 & 0.393 & 0.532 \\
\bottomrule
\end{tabular}
% }
\end{table}

\subsection{Ablation Study}

\subsubsection{Effectiveness of CT-Agent Planning}

To evaluate the effect of anatomical planning, we compare CT-Agent with a standard end-to-end baseline LLaVA-CT that lacks region-specific routing. As shown in Table~\ref{tab:report-comparison} and Table~\ref{tab:agent-comparison}, CT-Agent outperforms the baseline by 0.199 CE-F1 in report generation, and achieves a 0.057 gain in average F1 for presence and 0.139 gain  for abnormality QA.

Furthermore, our error analysis reveals that the General Model often suffers from anatomical confusion—e.g., mistaking mediastinum abnormalities as lung-related or confusing heart with pleura—due to the lack of region-guided control. In contrast, the CT-Agent architecture explicitly guides the model's attention to the correct region via a structured planning mechanism, effectively reducing such structural-level errors. These results demonstrate the advantage and interpretability of agent-based planning in 3D medical QA.

\subsubsection{Effectiveness of Hierarchical Token Compression}
We conducted ablation studies to evaluate the efficacy of the token compression module in the CT-Agent framework by systematically replacing its components with four alternative token processing strategies. First, we substituted the original module with a simple truncation strategy that discards excess tokens. Second, we implemented slice-level random sampling to reduce sequence length. Third, fixed slice sampling was tested by selecting slices at predetermined intervals. Fourth, we deployed our proposed compression module without global tokens, and finally, integrated the complete module with global tokens.
As shown in Table~\ref{tab:compression-comparison}, truncation significantly degraded performance in lung-related tasks, with precision and recall in lung presence detection dropping to 0.857 and 0.319, respectively, due to the loss of slice information. Random sampling caused instability, yielding only 0.505 F1 in lung presence detection, while fixed-interval sampling offered better consistency but was limited to anatomical variations. CT-Agent without global tokens reduced performance slightly, from 0.566 to 0.559 in lung abnormality F1. The full CT-Agent achieved the best results, with 0.941 recall and 0.952 F1 in lung presence detection, showing that adaptive compression and global tokens enhance local feature retention and cross-slice reasoning. The performance gap emphasizes our method's value for 3D CT visual question answering. Similar trends were observed in heart-related metrics, further confirming the effectiveness of our design.

\begin{table}[t]
\centering
\caption{Comparison of token compression strategies in CT-Agent, including truncation, random/fixed sampling, and variants with or without global tokens.}
\label{tab:compression-comparison}
\scriptsize
\setlength{\tabcolsep}{6pt}
\renewcommand{\arraystretch}{1.1}
% \begin{tabular}{c|ccc|ccc|ccc|ccc}
\begin{tabular}{
    >{\centering\arraybackslash}p{3.55 cm} |
    >{\centering\arraybackslash}p{0.59cm}
    >{\centering\arraybackslash}p{0.59cm}  
    >{\centering\arraybackslash}p{0.66cm} |
    >{\centering\arraybackslash}p{0.59cm} 
    >{\centering\arraybackslash}p{0.59cm}
    >{\centering\arraybackslash}p{0.66cm} |
    >{\centering\arraybackslash}p{0.59cm}
    >{\centering\arraybackslash}p{0.59cm} 
    >{\centering\arraybackslash}p{0.66cm} |
    >{\centering\arraybackslash}p{0.59cm} 
    >{\centering\arraybackslash}p{0.59cm} 
    >{\centering\arraybackslash}p{0.66cm} 
}
\toprule
\multirow{3}{*}{\textbf{Method}} 
& \multicolumn{6}{c}{\textbf{Lung}} 
& \multicolumn{6}{c}{\textbf{Heart}} \\
\cmidrule(lr){2-7} \cmidrule(lr){8-13}
& \multicolumn{3}{c}{\textbf{Presence}} & \multicolumn{3}{c}{\textbf{Abnormality}} 
& \multicolumn{3}{c}{\textbf{Presence}} & \multicolumn{3}{c}{\textbf{Abnormality}} \\
\cmidrule(lr){2-4} \cmidrule(lr){5-7} \cmidrule(lr){8-10} \cmidrule(lr){11-13}
& \textbf{P} & \textbf{R} & \textbf{F1} 
& \textbf{P} & \textbf{R} & \textbf{F1} 
& \textbf{P} & \textbf{R} & \textbf{F1} 
& \textbf{P} & \textbf{R} & \textbf{F1} \\
\midrule
No Compression (trunc.) 
& 0.857 & 0.319 & 0.465 & 0.835 & 0.141 & 0.242 
& 0.810 & 0.354 & 0.493 & 0.786 & 0.289 & 0.418 \\

Random Sampling 
& 0.894 & 0.352 & 0.505 & 0.882 & 0.251 & 0.391 
& 0.850 & 0.556 & 0.672 & 0.821 & 0.460 & 0.590 \\

Fixed-interval Sampling 
& 0.862 & 0.375 & 0.523 & 0.818 & 0.282 & 0.419 
& 0.841 & 0.713 & 0.772 & 0.802 & 0.650 & 0.718 \\

CT-Agent (w/o Global) 
& \textbf{0.970} & 0.841 & 0.901 & \textbf{0.932} & 0.399 & 0.559 
& \textbf{0.876} & 0.835 & 0.855 & 0.834 & 0.726 & 0.776 \\

CT-Agent (w/ Global) 
& 0.962 & \textbf{0.941} & \textbf{0.952} 
& 0.917 & \textbf{0.410} & \textbf{0.566} 
& 0.868 & \textbf{0.894} & \textbf{0.880} 
& \textbf{0.842} & \textbf{0.728} & \textbf{0.781} \\
\bottomrule
\end{tabular}
\end{table}

\subsubsection{Effectiveness of Prediction-Guided Exemplar Retrieval}

To evaluate the effectiveness of the prediction-guided exemplar retrieval module, we compare it with two alternative strategies: (1) zero-shot baseline, where no exemplar is provided and the model solely relies on its internal knowledge to generate reports. (2) static few-shot baseline, where a fixed set of exemplars is used for all inputs without semantic matching.

As shown in Table \ref{tab:rag-comparison}, replacing prediction-guided retrieval with static few-shot reduces BLEU-4 and METEOR by 0.021 and 0.026, respectively, while zero-shot substitution amplifies these declines to 0.095 and 0.090. The sharp drops in BLEU-3/4 highlight degraded fluency and semantic coherence, and the synchronized METEOR decline reveals reduced lexical diversity and semantic precision.

We observe that static few-shot lacks input-specific adaptability, and zero-shot strategies over-rely on model generalization without contextual guidance. In contrast, prediction-guided retrieval dynamically selects semantically aligned examples via intermediate predictions, resolving these limitations. This mechanism significantly improves report quality, confirming its necessity for robust and context-aware natural language generation in CT-Agent.

\begin{table}[t]
\centering
\caption{Comparison of different exemplar retrieval strategies on radiology report generation. We report standard NLG metrics: BLEU-1 to BLUE-4 (BL-1 to BL-4), ROUGE-L (RL), and METEOR (M).
CT-Agent (w/ Zero) uses no exemplars (zero-shot), CT-Agent (w/ Stat) uses statically selected exemplars, and 
CT-Agent (w/ Retr) uses prediction-guided exemplar retrieval.
}
\label{tab:rag-comparison}
\begin{tabular}{
    >{\centering\arraybackslash}p{3.78cm} |
    >{\centering\arraybackslash}p{1.62cm}
    >{\centering\arraybackslash}p{1.62cm}
    >{\centering\arraybackslash}p{1.62cm}
    >{\centering\arraybackslash}p{1.62cm}
    >{\centering\arraybackslash}p{1.62cm}
    >{\centering\arraybackslash}p{1.62cm}
}
\toprule
\textbf{Method} & \textbf{BL-1} & \textbf{BL-2} & \textbf{BL-3} & \textbf{BL-4} & \textbf{RL} & \textbf{M} \\
\midrule
CT-Agent (w/ Zero) & 0.423 & 0.274 & 0.188 & 0.136 & 0.346 & 0.335 \\
CT-Agent (w/ Stat) & 0.484 & 0.349 & 0.265 & 0.210 & 0.440 & 0.399 \\
CT-Agent (w/ Retr) & \textbf{0.502} & \textbf{0.374} & \textbf{0.290} & \textbf{0.231} & \textbf{0.490} & \textbf{0.425} \\
\bottomrule
\end{tabular}
\end{table}

Moreover, detailed qualitative analysis reveals that static few-shot prompting often leads to irrelevant or suboptimal exemplar usage, negatively affecting report coherence and accuracy. In contrast, our dynamic retrieval method effectively aligns exemplar context with the specific query, substantially enhancing the overall quality and clinical relevance of the generated reports.

\section{Conclusion}

In this paper, we presented CT-Agent, a multimodal-LLM agent designed for 3D CT radiology question answering. 
CT-Agent addresses two fundamental challenges of 3D CT analysis, anatomical complexity and spatial relationship through an anatomy-aware and a global-token compression strategy. CT-Agent enables interpretable, region-specific inference while preserving token efficiency and semantic integrity.
Extensive experiments on CT-RATE and RadGenome-ChestCT datasets demonstrate that CT-Agent outperforms state-of-the-art methods in both natural language generation and clinical efficacy. Our ablation studies further validate the effectiveness of each core component.
Notwithstanding the demonstrated effectiveness of CT-Agent in both natural language generation and clinical efficacy on 3D chest CT tasks, there remain several limitations. In future work, we will integrate multi-modal clinical evidence, such as prior radiology reports, to enhance contextual reasoning. Moreover, we plan to incorporate longitudinal scan analysis and real-time physician feedback to support interactive and temporally-aware diagnostic workflows.

\bibliography{ref}

\appendix
\section{Prompt Template}

\begin{lstlisting}
Task Classification Template:
You are a medical-domain assistant that classifies the user's intent and extracts anatomical focus from the question if applicable. 

Given a natural language query related to a chest CT scan, your task is to:
- Determine whether the query is a radiology report generation request or a region-guided question(QA task).
- If it is a QA task, identify the anatomical region(s) mentioned or implied in the question.
   Choose from the following predefined regions:
   ["Trachea and Bronchi", "Thyroid", "Lung", "Heart", "Mediastinum", "Pleura", "Esophagus", "Abdomen", "Bone", "Breast"]
- If it is a report generation task, leave the `target_region` as an empty list.

Return your results in the following JSON format:
```json
{
  "task_type": "QA" or "Report",
  "target_region": [list of anatomical regions or empty list]
}

User Query: {{user_question}}
\end{lstlisting}

\begin{lstlisting}
Query Rewriting Template:

You are a clinical assistant helping standardize diagnostic questions for chest CT interpretation. Given:
- A user question written in free-form natural language.
- A predefined set of clinical question templates.
- The anatomical region identified by the system.

Your task:
- Identify the clinical intent of the user question (e.g., presence detection, abnormality localization, or size estimation).
- Choose the most appropriate predefined template.
- Fill in the placeholders (e.g., {region}, {abnormality}) using information inferred from the user's question.
- If the abnormality type is not explicitly mentioned, use a general placeholder like "abnormality".

Predefined Question Templates:
1. What are the abnormalities in the {region}?
2. What is the approximate size of the {abnormality} in the {region}?
3. Where is the {abnormality} located in the image?
4. Can {abnormality} be identified in the {region}?

Input:
User Question: {{user_question}}
Target Anatomical Region: {{region}}

Output Format:
Rewritten Clinical Query: {{generated_question}}
\end{lstlisting}

\begin{lstlisting}
Answer Generation Template:

You are a medical-domain assistant. Your task is to answer the user's original question by referencing the provided professional version of the question and its corresponding answer. Assume the reference question is a clinically accurate reformulation of the user's intent. Use its answer as the basis for your response. If the reference answer does not fully cover the user's question, you may cautiously infer missing details based on established medical knowledge, and explain your reasoning briefly.

User Question: {{user_question}}

Reference question: {{reference_question}}

Reference Answer: {{reference_answer}}

Output Format: Answer: {{generated answer}}

\end{lstlisting}

\begin{lstlisting}
Report Generation Template:

You are a board-certified radiologist. Given structured findings for each anatomical region of a chest CT scan, generate a clinically styled radiology report that matches expert-written radiology report style and language patterns.

Follow these detailed instructions:

Anatomical Regions:
Report in the following strict order:
Trachea and Bronchi > Thyroid > Lung > Heart > Mediastinum > Pleura > Esophagus > Abdomen > Bone > Breast

Language Style and Formatting:
- Use passive voice, objective tone, and formal clinical phrasing.
- For each region, begin with a normal finding sentence if applicable.
- For abnormal findings, follow the 4-part structure:
  [Anatomical location] + [Image Finding] + [Measurement if applicable] + [Interpretation]


Standard Sentence Templates:

Normal Description Template (use exactly when applicable):
- "Trachea and both main bronchi are open."
- "No occlusive pathology was detected in the trachea and both main bronchi."
- "Heart contour and size are normal."
- "Pericardial effusion-thickening was not observed."
- "Thoracic esophagus calibration was normal and no significant wall thickening was detected."
- "Mediastinal main vascular structures, heart contour, size are normal."
- "No enlarged lymph nodes in pathological dimensions were detected."
- "Pleural effusion-thickening was not detected."
- "No space-occupying lesion was detected in the liver that entered the cross-sectional area."
- "Bone structures in the study area are natural."
- "Vertebral corpus heights are preserved."

Abnormal Description Patterns (apply when findings exist):
Use phrases like:
- "Ground-glass opacities are observed in the [lung region], especially in the [peripheral areas]."
- "Millimetric nodules are observed in both lungs, the largest measuring [X] mm in [location]."
- "Subsegmental atelectasis areas are noted in the [segment]."
- "There is a pleural effusion with loculation measuring [X] cm at its thickest point."
- "A [X] mm diameter calculus was observed in the gallbladder lumen."
- "Diffuse degenerative changes and osteophytic taperings are noted in the thoracic vertebrae."

Interpretation Language:
- "Findings are evaluated in favor of [diagnosis]."
- "Findings appear stable."
- "It is recommended to be evaluated together with clinical and laboratory results."
- "No mass lesion with distinguishable borders was detected."
- "As far as can be observed..."

Final Output:
- Combine all findings into a fluent, multi-paragraph report, organized in the specified order.
- Avoid subjective judgments or treatment suggestions.
- Do not summarize or conclude; only report imaging findings.
- Use radiological English close to examples above to improve quality and consistency.
- Try to use longer and more coherent sentences as much as possible.

Input structured finding: {{inputs}}

Examples: {{examples}}
\end{lstlisting}

\end{document}